\documentclass[10pt,twocolumn,letterpaper]{article}

\usepackage{cvpr}              

\usepackage{amsmath}
\usepackage{amssymb}
\usepackage{subcaption}

\usepackage{graphicx}

\makeatletter
\@namedef{ver@everyshi.sty}{}
\makeatother
\usepackage{tikz}

\usepackage{comment}
\usepackage{color}
\usepackage{sidecap}

\usepackage[accsupp]{axessibility}  

\usepackage{booktabs}
\usepackage{mdwlist}

\graphicspath{{Figures/}}
\usepackage{caption}
\usepackage{enumitem}
\setitemize{noitemsep,topsep=0pt,parsep=0pt,partopsep=0pt}
\usepackage{newfloat}
\usepackage{textcomp,multirow,upgreek}
\usepackage[pagebackref=true,breaklinks=true,letterpaper=true,colorlinks,bookmarks=false]{hyperref}

\usepackage[capitalize]{cleveref}
\crefname{section}{Sec.}{Secs.}
\Crefname{section}{Section}{Sections}
\Crefname{table}{Table}{Tables}
\crefname{table}{Tab.}{Tabs.}

\def\cvprPaperID{7732} 
\def\confName{CVPR}
\def\confYear{2023}

\begin{document}

\title{Graph Transformer GANs for Graph-Constrained House Generation}

\author{
Hao Tang$^1$ \quad Zhenyu Zhang$^2$ \quad Humphrey Shi$^3$ \quad Bo Li$^2$ \\
Ling Shao$^4$ \quad   Nicu Sebe$^5$ \quad Radu Timofte$^{1,6}$ \quad Luc Van Gool$^{1,7}$ \\
$^1$CVL, ETH Zurich \quad $^2$Tencent Youtu Lab \quad $^3$U of Oregon \& UIUC \& Picsart AI Research \\
$^4$UCAS-Terminus AI Lab, UCAS \quad $^5$University of Trento \quad$^6$University of Wurzburg \quad $^7$KU Leuven
}
\maketitle

\begin{abstract}
	We present a novel graph Transformer generative adversarial network (GTGAN) to learn effective graph node relations in an end-to-end fashion for the challenging graph-constrained house generation task. The proposed graph-Transformer-based generator includes a novel graph Transformer encoder that combines graph convolutions and self-attentions in a Transformer to model both local and global interactions across connected and non-connected graph nodes.
	Specifically, the proposed connected node attention (CNA) and non-connected node attention (NNA) aim to capture the global relations across connected nodes and non-connected nodes in the input graph, respectively. The proposed graph modeling block (GMB) aims to exploit local vertex interactions based on a house layout topology. Moreover, we propose a new node classification-based discriminator to preserve the high-level semantic and discriminative node features for different house components. Finally,  we propose a novel graph-based cycle-consistency loss that aims at maintaining the relative spatial relationships between ground truth and predicted graphs.
	Experiments on two challenging graph-constrained house generation tasks (\ie, house layout and roof generation) with two public datasets demonstrate the effectiveness of GTGAN in terms of objective quantitative scores and subjective visual realism. New state-of-the-art results are established by large margins on both tasks.
\end{abstract}

\section{Introduction}

\begin{figure*}[!t] \small
	\centering
	\includegraphics[width=0.95\linewidth]{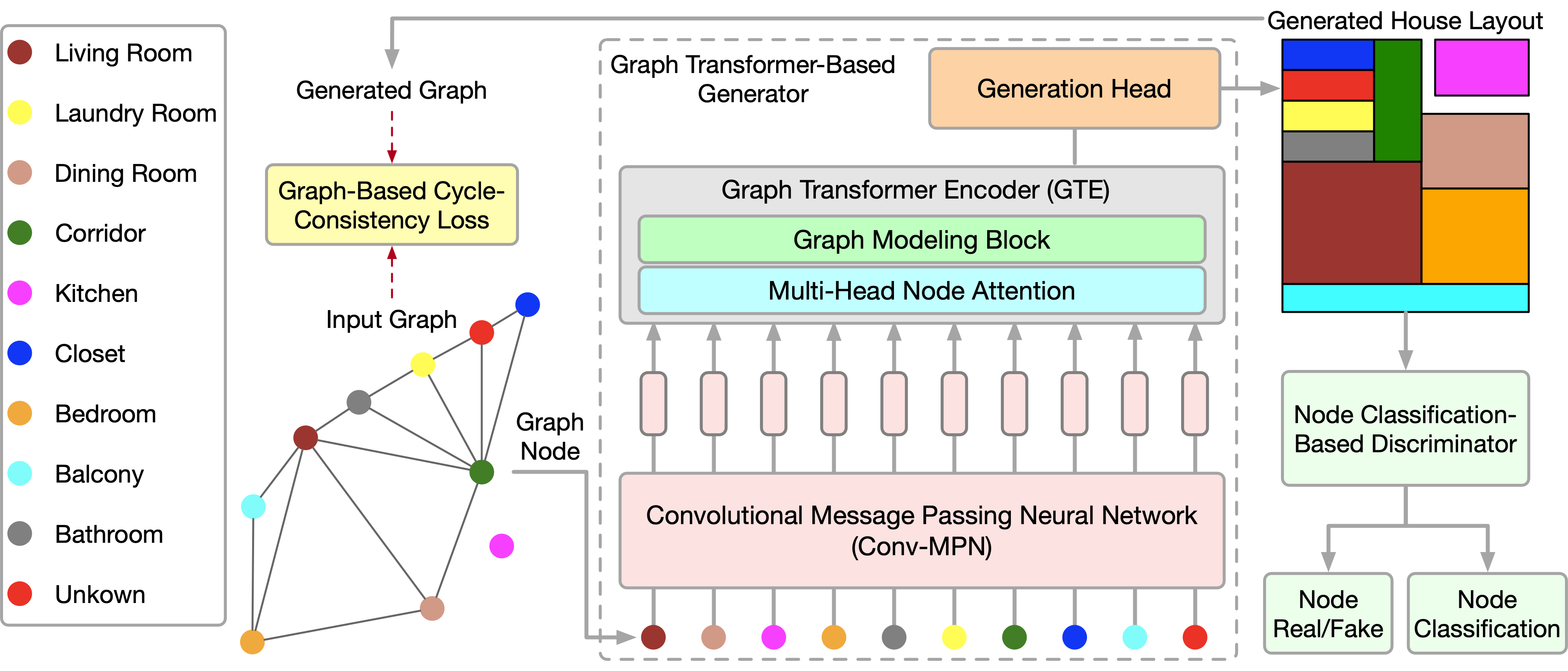}
	\caption{Overview of the proposed GTGAN on house layout generation. It consists of a novel graph Transformer-based generator $G$ and a novel node classification-based discriminator $D$. 
		The generator takes graph nodes as input and aims to capture local and global relations across connected and non-connected nodes using the proposed graph modeling block and multi-head node attention, respectively.
		Note that we do not use position embeddings since our goal is to predict positional node information in the generated house layout.
		The discriminator $D$ aims to distinguish real and generated layouts and simultaneously classify the generated house layouts to their corresponding room types.
		The graph-based cycle-consistency loss aligns the relative spatial relationships between ground truth and predicted nodes.
		The whole framework is trained in an end-to-end fashion so that all components benefit from each other.}
	\label{fig:framework}
	\vspace{-0.4cm}
\end{figure*}

\sloppy
This paper focuses on converting an input graph to a realistic house footprint, as depicted in Figure~\ref{fig:framework}.
Existing house generation methods such as \cite{wang2019planit,hu2020graph2plan,ashual2019specifying,johnson2018image,nauata2020house,wu2020pq,qian2021roof}, typically rely on building convolutional layers. 
However, convolutional architectures lack an understanding of long-range dependencies in the input graph since inherent inductive biases exist.
Several Transformer architectures \cite{vaswani2017attention,dosovitskiy2020image,zheng2020rethinking,wang2020max,wang2021transbts,carion2020end,zhu2020deformable,huang2020hand,huang2020hot,lin2020end,chen2020topological} based on the self-attention mechanism have recently been proposed to encode long-range or global relations, thus learn highly expressive feature representations.
On the other hand, graph convolution networks are good at exploiting local and neighborhood vertex correlations based on a graph topology.
Therefore, it stands to reason to combine graph convolution networks and Transformers to model local as well as global interactions for solving graph-constrained house generation.

To this end, we propose a novel graph Transformer generative adversarial network (GTGAN), which consists of two main novel components, \ie, a graph Transformer-based generator and a node classification-based discriminator (see Figure~\ref{fig:framework}).
The proposed generator aims to generate a realistic house from the input graph, which consists of three components, \ie, a convolutional message passing neural network (Conv-MPN), a graph Transformer encoder (GTE), and a generation head.
Specifically, Conv-MPN first receives graph nodes as inputs and aims to extract discriminative node features.
Next, the embedded nodes are fed to GTE, in which the long-range and global relation reasoning is performed by the connected node attention (CNA) and non-connected node attention (NNA) modules. 
Then, the output from both attention modules is fed to the proposed graph modeling block (GMB) to capture local and neighborhood relationships based on a house layout topology.
Finally, the output of GTE is fed to the generative head to produce the corresponding house layout or roof.
To the best of our knowledge, we are the first to use a graph Transformer to model local and global relations across graph nodes for solving graph-constrained house generation.

In addition, the proposed discriminator aims to distinguish
real and fake house layouts, which ensures that our generated house layouts or roofs look realistic.
At the same time, the discriminator classifies the generated rooms to their corresponding real labels, preserving the discriminative and semantic features (\eg, size and position) for different house components.
To maintain the graph-level layout, we also propose a novel graph-based cycle-consistency loss to preserve the relative spatial relationships between ground truth and predicted graphs.

Overall, our contributions are summarized as follows:

\begin{itemize}[leftmargin=*]
	\item We propose a novel Transformer-based network (\ie, GTGAN) for the challenging graph-constrained house generation task. 
	To the best of our knowledge, GTGAN is the first Transformer-based framework, enabling more effective relation reasoning for composing house layouts and validating adjacency constraints.
	\item We propose a novel graph Transformer generator that combines both graph convolutional networks and Transformers to explicitly model global and local correlations across both connected and non-connected nodes simultaneously.
	We also propose a new node classification-based discriminator to preserve high-level semantic and discriminative features for different types of rooms.
	\item We propose a novel graph-based cycle-consistency loss to guide the learning process toward accurate relative spatial distance of graph nodes.
	\item Qualitative and quantitative experiments on two challenging graph-constrained house generation tasks (\ie, house layout generation and house roof generation) with two datasets demonstrate that GTGAN can generate better house structures than state-of-the-art methods, such as HouseGAN~\cite{nauata2020house} and RoofGAN \cite{qian2021roof}.
\end{itemize}

\section{Related Work}

\noindent \textbf{Generative Adversarial Networks}
\cite{goodfellow2014generative} have been widely used for image generation \cite{karras2018style,shaham2019singan,tang2022local,tang2022multi}.
The vanilla GAN consists of a generator and a discriminator. 
The generator aims to synthesize photorealistic images from a noise vector, while the discriminator aims to distinguish between real and generated samples.
To create user-specific images, the conditional GAN (CGAN) \cite{mirza2014conditional} was proposed.
A CGAN combines a vanilla GAN and external information, such as class labels~\cite{choi2017stargan}, text descriptions \cite{han2017stackgan,tao2023galip,tao2022df}, object keypoints~\cite{reed2016learning}, human skeletons~\cite{tang2020xinggan}, semantic maps~\cite{tang2019multi,park2019semantic,tang2020local}, edge maps \cite{tang2023edge}, or attention maps~\cite{mejjati2018unsupervised}.
This paper mainly focuses on the challenging graph-constrained generation task, which aims to transfer an input graph to a realistic house.

\noindent \textbf{Graph-Constrained Layout Generation}
has been a focus of research recently \cite{dhamo2021graph,luo2020end,wang2019planit,hu2020graph2plan}. 
For example, Wang et al. \cite{wang2019planit} presented a layout generation framework that plans an indoor scene as a relation graph and iteratively inserts a 3D model at each node.
Hu et al. \cite{hu2020graph2plan} converted a layout graph along with a building boundary into a floorplan that fulfills both the layout and boundary constraints. 
Ashual et al.~\cite{ashual2019specifying} and Johnson et al.~\cite{johnson2018image} tried to generate image layouts and synthesize realistic images from input scene graphs via GCNs. 
Nauata et al.~\cite{nauata2020house} proposed a graph-constrained generative
adversarial network, whose generator and discriminator are built upon
a relational architecture.
Our innovation is a novel graph Transformer GAN, where the input constraint is
encoded into the graph structure of the proposed graph Transformer-based generator and node classification-based discriminator.
Experimental results show the effectiveness of GTGAN over all the leading methods.

\noindent \textbf{Transformers in Computer Vision.}
The Transformer was first proposed in \cite{vaswani2017attention} for machine translation
and has established state-of-the-art results in many natural language processing (NLP) tasks. 
Recently, the Vision Transformer (ViT) \cite{dosovitskiy2020image} equipped with global self-attention has achieved state-of-the-art results on the classification task.
Since then, Transformer-based approaches have been shown to be efficient in many computer vision tasks including image segmentation \cite{zheng2020rethinking,wang2020max,wang2021transbts}, object detection \cite{carion2020end,zhu2020deformable,dai2022ao2}, depth estimation \cite{yang2021transformer}, pose estimation \cite{li2022mhformer,lin2020end}, video inpainting \cite{zeng2020learning}, vision-and-language navigation \cite{chen2020topological}, video classification \cite{neimark2021video}, human reaction generation \cite{chopin2023interaction}, 3D pose transfer \cite{chen2022geometry,chen2021aniformer}.
Different from these methods, in this paper, we adopt a Transformer-based network to tackle the graph-constrained house generation task.
However, integrating graph convolutional networks and ViTs is not trivial.
To this end, we propose a graph Transformer-based generator to capture both local and global relations across nodes in a graph.
To the best of our knowledge, GTGAN is the first Transformer-based house generation framework.

\section{The Proposed Graph Transformer GAN}

This section presents the details of the proposed GTGAN, which consists of a novel graph Transformer-based generator $G$, a node classification-based discriminator $D$, and a graph-based cycle-consistency loss. 
An illustration of the proposed GTGAN framework is shown in Figure~\ref{fig:framework}.

\subsection{Graph Transformer-Based Generator}
We only illustrate the details of our contributions on the house layout generation task for simplicity. The extension of the proposed contributions to house roof generation is straightforward.
Take house layout generation in Figure~\ref{fig:framework} as an example. The generator $G$ receives a noise vector for each room and a bubble diagram as inputs. It then generates a house layout, where each room is represented as an axis-aligned rectangle. 
We represent each bubble diagram as a graph, where each node represents a room of a certain type, and each edge represents the spatial adjacency. 
Specifically, we generate a rectangle for each room, where two rooms with a graph edge should be spatially adjacent, while two rooms without an edge should be spatially dis-adjacent. 

\noindent \textbf{Input Graph Representation.}
Given a bubble diagram, we first generate a node for each room and initialize it with a $128$-d noise vector sampled from a normal distribution.
We then concatenate the noise vector with a $10$-d room type vector $\overrightarrow{t_r}$ ($r$ is a room index), encoded in the one-hot format.
Therefore, we can obtain a $138$-d vector $\overrightarrow{g_r}$ to represent the input bubble diagram as follows,
\begin{equation}
	\begin{aligned}
		\overrightarrow{g_r} \leftarrow \left \{ \mathbb{N}(0, 1)^{128}; \overrightarrow{t_r} \right \}.
	\end{aligned}\label{eq:equation-factor}
\end{equation}
Note that, different from the highly successful ViT \cite{dosovitskiy2020image}, we use graph nodes as the input of the proposed graph Transformer instead of using image patches, which makes our framework very different.

\noindent \textbf{Convolutional Message Passing Neural Network.}
As indicated in HouseGAN \cite{nauata2020house}, Conv-MPN stores feature as a 3D tensor in the output design space. 
We thus apply a shared linear layer to expand $\overrightarrow{g_r}$ into a feature volume ${\rm \bf{g}}_r^{l=1}$ of size $16 {\times} 8 {\times} 8$, where $l {=} 1$ is the feature extracted from the first Conv-MPN layer, which will be upsampled twice using a transposed convolution
to become a feature volume ${\rm \bf{g}}_r^{l=3}$ of size $16 {\times} 32 {\times} 32$.

The Conv-MPN layer updates a graph of room-wise feature volumes via a convolutional message passing \cite{zhang2020conv}.
Specifically, we update ${\rm \bf{g}}_r^{l=1}$ over the following steps: 
1) We use a GTE to capture the long-range correlations across rooms that are connected in the input graph;
2) We employ another GTE to capture the long-range dependencies across non-connected rooms in the input graph;
3) We concatenate a sum-pooled feature across connected rooms in the input graph; 
4) We concatenate a sum-pooled feature across non-connected rooms; 
and 
5) We apply a convolutional neural network (CNN) on the combined feature.
This process can be formulated as follows,
\begin{equation}
	\begin{aligned}
		{\rm \bf{g}}_r^l \leftarrow {\rm CNN}\left[ {\rm \bf{g}}_r^l + {\rm GTE}\left( \underset{s \in {\rm N}(r)}{\rm Pool}  {\rm \bf{g}}_s^l, {\rm \bf{g}}_r^l \right) + 
		\right. \\ \phantom{=\;\;}\left.  {\rm GTE}\left( \underset{s \in {\rm \overline{N}}(r)}{\rm Pool}  {\rm \bf{g}}_s^l, {\rm \bf{g}}_r^l \right);   \underset{s \in {\rm N}(r)}{\rm Pool}  {\rm \bf{g}}_s^l;   \underset{s \in {\rm \overline{N}}(r)}{\rm Pool}  {\rm \bf{g}}_s^l \right],
	\end{aligned}\label{eq:eq}
\end{equation}
where ${\rm {N}}(r)$ and ${\rm \overline{N}}(r)$ denote sets of rooms that are connected and not-connected, respectively;
``+'' and ``;'' denote pixel-wise addition and channel-wise concatenation, respectively.
We also explore two more variations (Eq.~\eqref{eq:eq1} and~\eqref{eq:eq2}) to validate  the effectiveness of  Eq.~\eqref{eq:eq} as follows,
\begin{equation}
	\begin{aligned}
		{\rm \bf{g}}_r^l  \leftarrow {\rm CNN}\left[ {\rm GTE}\left( \underset{s \in {\rm N}(r)}{\rm Pool}  {\rm \bf{g}}_s^l, {\rm \bf{g}}_r^l \right) + \right. \\ \phantom{=\;\;}\left.  {\rm GTE}\left( \underset{s \in {\rm \overline{N}}(r)}{\rm Pool}  {\rm \bf{g}}_s^l, {\rm \bf{g}}_r^l \right);   \underset{s \in {\rm N}(r)}{\rm Pool}  {\rm \bf{g}}_s^l;   \underset{s \in {\rm \overline{N}}(r)}{\rm Pool}  {\rm \bf{g}}_s^l \right].
		\label{eq:eq1}
	\end{aligned}
\end{equation}
\begin{equation}
	\begin{aligned}
		{\rm \bf{g}}_r^l  \leftarrow {\rm CNN}\left[ {\rm \bf{g}}_r^l +  {\rm GTE}\left( \underset{s \in {\rm N}(r)}{\rm Pool}  {\rm \bf{g}}_s^l, {\rm \bf{g}}_r^l \right) +  \right. \\ \phantom{=\;\;}\left. {\rm GTE}\left( \underset{s \in {\rm \overline{N}}(r)}{\rm Pool}  {\rm \bf{g}}_s^l, {\rm \bf{g}}_r^l \right)\right].
		\label{eq:eq2}
	\end{aligned}
\end{equation}

\begin{figure}[!t] \small
	\centering
	\includegraphics[width=1\linewidth]{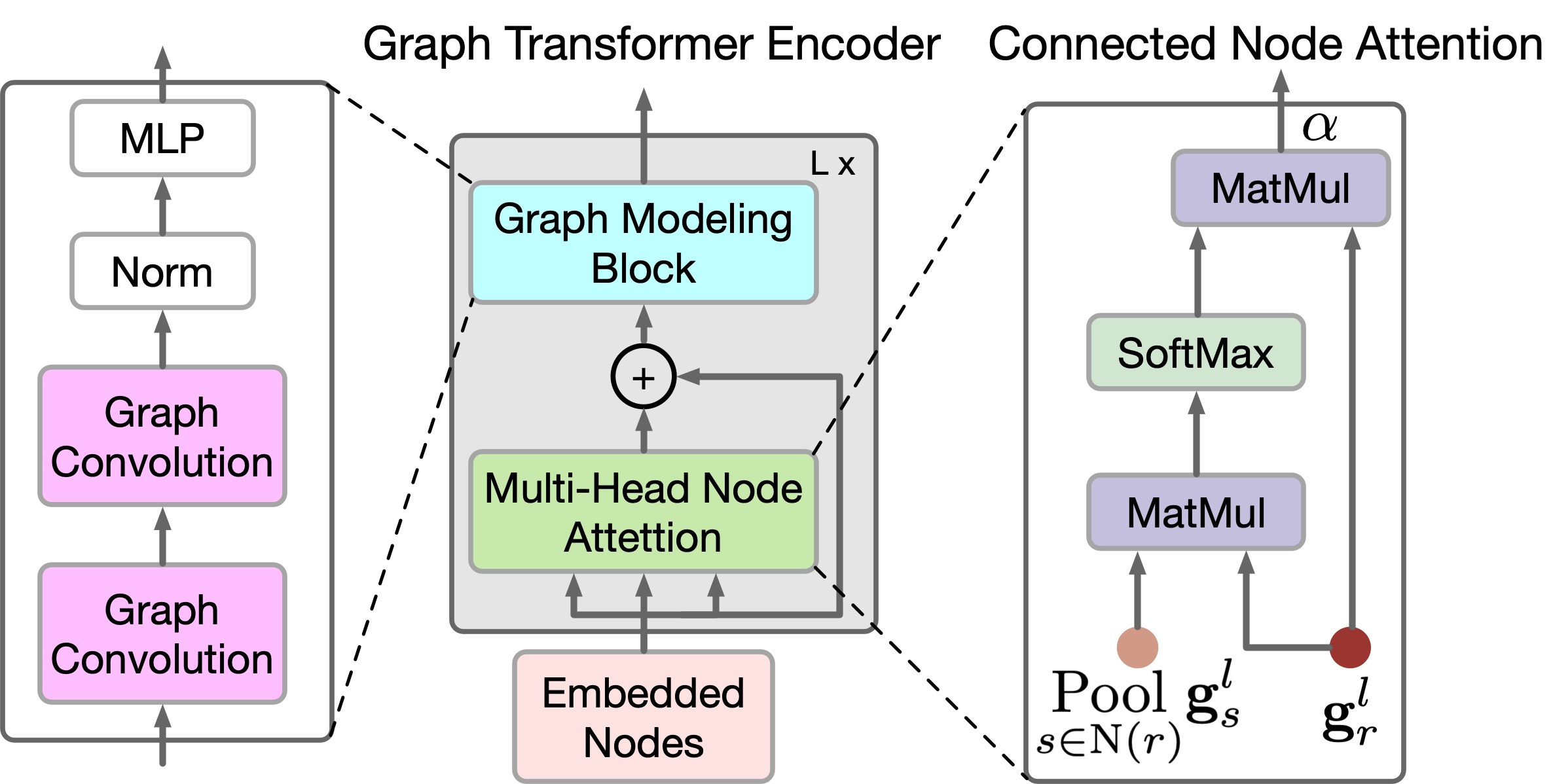}
	\caption{Overview of the proposed graph Transformer encoder, which consists of a multi-head node attention and a graph modeling block. It can capture both global and local correlations for graph-constrained house generation. This encoder consists of $L{=}8$ identical blocks. The proposed connected node attention aims to capture long-range relations across connected nodes.  Note that the proposed non-connected node attention has the same structure as the connected node attention but takes non-connected nodes as input.  It aims to capture long-range relations across non-connected nodes.}
	\label{fig:block}
	\vspace{-0.4cm}
\end{figure}

\noindent \textbf{Node Attentions in Graph Transformer Encoder.}
To capture local and global relationships across graph nodes, we propose a novel GTE, as shown in Figure~\ref{fig:block}.
GTE combines self-attention in Transformer and graph convolution networks to capture global and local correlations, respectively.
Note that we do not use position embeddings in our framework since our goal is to generate node positions in the generated house layout.

The proposed GTE is quite different from the one presented in ViT \cite{dosovitskiy2020image} since the two input modalities are different, \ie, images and graph nodes.
Thus, we extend the multi-head self-attention in \cite{dosovitskiy2020image} to the multi-head node attention, which aims to capture  the global correlations across connected rooms/nodes and the global dependencies across non-connected rooms/nodes.
To this end, we propose two novel graph node attention modules, \ie, connected node attention (CNA) and non-connected node attention (NNA).
Both CNA and NNA share the same network structure.

The goal of CNA (see Figure~\ref{fig:block}) is to model the global correlations across connected rooms.
Specifically, we perform a matrix multiplication between the transpose of $\underset{s \in {\rm N}(r)}{\rm Pool} {\rm \bf{g}}_s^l$ and ${\rm \bf{g}}_r^l$, and apply a Softmax function to calculate the connected node attention map $\underset{{\rm N}(r)}{\rm Att}$,
\begin{equation}
	\begin{aligned}
		\underset{{\rm N}(r)}{\rm Att} = {\rm softmax} \Bigg [ \frac{{\rm \bf{g}}_r^l \Big(\underset{s \in {\rm N}(r)}{\rm Pool} {\rm \bf{g}}_s^l \Big)^\top} {\sqrt{card(N(r))}} \Bigg],
		\label{eq:non_local1}
	\end{aligned}
\end{equation}
where $card(N(r))$ is the number of connected graph nodes in a training batch.
The connected node attention map $\underset{{\rm N}(r)}{\rm Att}$ measures the connected node's impact on other connected nodes. 
Then we perform matrix multiplication between ${\rm \bf{g}}_r^l$ and the transpose of $\underset{{\rm N}(r)}{\rm Att}$. 
Lastly, we multiply the result by a scaling parameter $\alpha$ to obtain the output,
\begin{equation}
	\begin{aligned}
		{\rm GTE}\left( \underset{s \in {\rm N}(r)}{\rm Pool}  {\rm \bf{g}}_s^l, {\rm \bf{g}}_r^l \right) = \alpha \sum_{1}^{N}\left (\underset{{\rm N}(r)}{\rm Att} \cdot {\rm \bf{g}}_r^l \right),
	\end{aligned}
\end{equation}
where $\alpha$ is a learnable parameter, initialized to 0, and learned by the model \cite{zhang2019self}.
By doing so, each connected node in ${\rm N}(r)$ is a weighted sum of all the connected nodes.
Thus, CNA obtains a global view of the spatial graph structure and can selectively adjust rooms according to the connected attention map, improving the house layout's representations and high-level semantic consistency.

Similarly, the goal of NNA is to capture global relations across non-connected rooms.
Specifically, we perform a matrix multiplication between the transpose of $\underset{s \in {\rm \overline{N}}(r)}{\rm Pool}  {\rm \bf{g}}_s^l$ and ${\rm \bf{g}}_r^l$, and apply a Softmax function to calculate the non-connected node attention map $\underset{{\rm \overline{N}}(r)}{\rm Att}$:

\begin{equation}
	\begin{aligned}
		\underset{{\rm \overline{N}}(r)}{\rm Att} = {\rm softmax} \Bigg[ \frac{{\rm \bf{g}}_r^l \Big(\underset{s \in {\rm \overline{N}}(r)}{\rm Pool} {\rm \bf{g}}_s^l \Big)^\top} {\sqrt{card(\overline{N}(r))}} \Bigg],
		\label{eq:non_local2}
	\end{aligned}
\end{equation}
where $card(\overline{N}(r))$ is the number of non-connected graph nodes in a training batch.
The non-connected node attention map $\underset{{\rm \overline{N}}(r)}{\rm Att}$ measures the non-connected  node's impact on other non-connected nodes. 
Then we perform matrix multiplication between ${\rm \bf{g}}_r^l$ and the transpose of $\underset{{\rm \overline{N}}(r)}{\rm Att}$. 
Lastly, we multiply the result by a scale parameter $\beta$ to obtain the output as follows,
\begin{equation}
	\begin{aligned}
		{\rm GTE}\left( \underset{s \in {\rm \overline{N}}(r)}{\rm Pool}  {\rm \bf{g}}_s^l, {\rm \bf{g}}_r^l \right) = 
		\beta \sum_{1}^{N}\left (\underset{{\rm \overline{N}}(r)}{\rm Att} \cdot {\rm \bf{g}}_r^l \right),
	\end{aligned}
\end{equation}
where $\beta$ gradually learns weights from 0 \cite{zhang2019self}.
By doing so, each non-connected node in ${\rm \overline{N}}(r)$ is a weighted sum of all the non-connected nodes.
Finally, we perform an element-wise sum with ${\rm \bf{g}}_r^l$ so that the updated node feature can capture both connected and non-connected spatial relations.
This process can be expressed as follows,
\begin{equation}
	\begin{aligned}
		{\rm \bf{g}}_r^l  \leftarrow {\rm \bf{g}}_r^l + {\rm GTE}\left( \underset{s \in {\rm N}(r)}{\rm Pool}  {\rm \bf{g}}_s^l, {\rm \bf{g}}_r^l \right) + {\rm GTE}\left( \underset{s \in {\rm \overline{N}}(r)}{\rm Pool}  {\rm \bf{g}}_s^l, {\rm \bf{g}}_r^l \right).
	\end{aligned}
	\label{eq:method2}
\end{equation}

\noindent \textbf{Graph Modeling in Graph Transformer Encoder.}
While CNA and NNA are useful for extracting long-range and global dependencies, it is less efficient at capturing fine-grained local information in complex house data structures. To fix this limitation, we propose a novel graph modeling block, as shown in Figure~\ref{fig:block}. 

Specifically, given the features ${\rm \bf{g}}_r^l $ generated in Eq. \eqref{eq:method2}, we further improve the local correlations by using graph convolutional networks as follows,
\begin{equation}
	\begin{aligned}
		{\rm \bf{\hat{g}}}_r^l = {\rm GC}(A, {\rm \bf{g}}_r^l; P) = \sigma(A {\rm \bf{g}}_r^l P),
	\end{aligned}
	\label{eq:graph}
\end{equation}
where $A$ denotes the adjacency matrix of a graph, ${\rm GC}(\cdot)$ represents graph convolution, and  $P$ the trainable parameters. 
$\sigma(\cdot)$ is the gaussian error linear unit (GeLU) proposed in \cite{hendrycks2016gaussian} activation function that aims to provide the network non-linearity.
We follow the structure design in GraphCMR \cite{kolotouros2019convolutional} to build our graph modeling block, which can explicitly encode the graph-constrained house structure within the network and thereby improve spatial locality in the feature representations.

\noindent \textbf{Generation Head.} 
We adopt three CNN layers to convert a feature volume into a room segmentation mask of size $1 {\times} 32 {\times} 32$. 
The numbers of convolutional channels are 256, 128, and 1, respectively.
We pass the graph of segmentation masks to the proposed discriminator $D$ during the training stage. 
Finally, we fit the tightest axis-aligned rectangle for each room to generate the house layout.

\subsection{Node Classification-Based Discriminator}
The input of the proposed discriminator is a graph of room segmentation masks, either from the generator or a real one.
The segmentation masks are of size $1 {\times} 32 {\times} 32$. 
We also take a $10$-d room type vector to preserve the room type information, and then we apply a linear layer to expand it to $8192$-d.
Next, we reshape it to a tensor of size $8 {\times} 32 {\times}32$.
Thus, we use a shared three-layer CNN to convert it to a feature of size $16 {\times} 32 {\times}32$, followed by two rounds of Conv-MPN and downsampling. 
Lastly, we use another three-layer CNN to convert each room feature into a $128$-d vector $\overrightarrow{d_r}$. 
To classify ground-truth samples from the generated ones, we sum-pool over all the room vectors and then apply a single linear layer to produce a scalar $\rm {\bf{\tilde{d}}_1}$, which can be expressed as follows,
\begin{equation}
	\begin{aligned}
		{\rm \bf{\tilde{d}}_1}  \leftarrow {\rm Linear} \left( \underset{r}{{\rm Pool}}~\overrightarrow{d_r} \right).
	\end{aligned}\label{eq:eq3}
\end{equation}
Moreover, we observe that HouseGAN \cite{nauata2020house} cannot produce very discriminative rooms, leading to similar generation results for different types of rooms.
To provide a more diverse generation for different rooms, we propose a novel node classification loss to learn more discriminative class-specific node representations.
Specifically, we sum pool over all the room vectors and add another single linear layer to output a $10$-d one-hot vector $\rm \bf{\tilde{d}}_2$, classifying generated rooms to the corresponding room labels,
\begin{equation}
	\begin{aligned}
		{\rm \bf{\tilde{d}}_2}  \leftarrow {\rm Linear} \left( \underset{r}{{\rm Pool}}~\overrightarrow{d_r} \right).
	\end{aligned}\label{eq:eq4}
\end{equation}
We use the binary cross-entropy loss between the real room label $\rm \bf{d}_2$ and the predicted label $\rm \bf{\tilde{d}}_2$.

\subsection{Graph-Based Cycle-Consistency Loss}
Providing global graph node relationship information is helpful in generating more accurate house layouts. 
To differentiate this process, we propose a novel loss based on an adjacency matrix that matches the spatial relationships between ground truth and generated graphs, as shown in Figure \ref{fig:framework}. 
Precisely, the graphs capture the adjacency relationships between each node of different rooms, and then we enforce the matching between the ground truth and generated graphs through the proposed graph-based cycle-consistency loss. 
Formally, we represent the graphs using two (square) weighted adjacency matrices of size $M{\times}M$,
\begin{equation}
	\begin{aligned}
		{\rm \textbf{G}}^{gt}  = \{g_{i,j}^{gt}\}_{\substack{i=1,\dots,M \\ j=1,\dots,M}},  \quad
		{\rm \textbf{G}}^{gen} = \{g_{i,j}^{gen}\}_{\substack{i=1,\dots,M \\ j=1,\dots,M}},
	\end{aligned}\label{eq:eq5}
\end{equation}
The matrix ${\rm \textbf{G}}^{gt}$ contains the adjacency information computed on ground
truth graph, while ${\rm \textbf{G}}^{gen}$ has the same information computed on the
generated graph. 
Note that we adapt the network in \cite{yu2019layout} to obtain the $G^{gen}$ from the generated house layout, followed by a fully-connected layer with the size of $M{\times}M$, then reshaped it to a square matrix.

In this way, each element of the matrices provides a measure of how close the two nodes $i$ and $j$ are in the ground truth and the generated graph, respectively. 
To measure the closeness between nodes, which is a hint of the strength of the connection between them, we consider weighted matrices where each entry $g_{i,j}$ depends on the shortest distance between them. 
For example, in the ground truth graph in Figure \ref{fig:framework}, the shortest distance between the dining room and the living room is 1, the shortest distance between the dining room and the closet is 2, and the shortest distance between the bedroom and the closet is 3.
Note that we do not consider self-connections, thus $g_{i,i}^{gt} {=} g_{i,i}^{gen} {=}0$ for $i{=}1,\dots,M$. 
Moreover, non-adjacent nodes have $-1$ as the entry. 
Then, we define the proposed graph-based cycle-consistency loss as the Frobenius norm between the two adjacency matrices,
\begin{equation}
	\mathcal{L}_{gcyc} = || {\rm \textbf{G}}^{gt} - {\rm \textbf{G}}^{gen} ||_F =|| {\rm \textbf{G}}^{gt} - G({\rm \textbf{G}}^{gt}) ||_F,
	\label{eq:eq6}
\end{equation}
where $G$ is the proposed graph Transformer-based generator. 
This loss function aims to faithfully maintain the reciprocal relationships between nodes. On the one hand, disjoint parts are enforced to be predicted as disjoint. 
On the other hand, neighboring nodes are enforced to be predicted as neighboring and to match the proximity ratios.

\section{Experiments}

The proposed GTGAN can be applied to different graph-based generative tasks such as house layout generation \cite{nauata2020house} and house roof generation \cite{qian2021roof}. In this section, we present experimental results and analysis on both tasks.

\begin{figure*}[!t] \small
	\centering
	\includegraphics[width=0.328\linewidth]{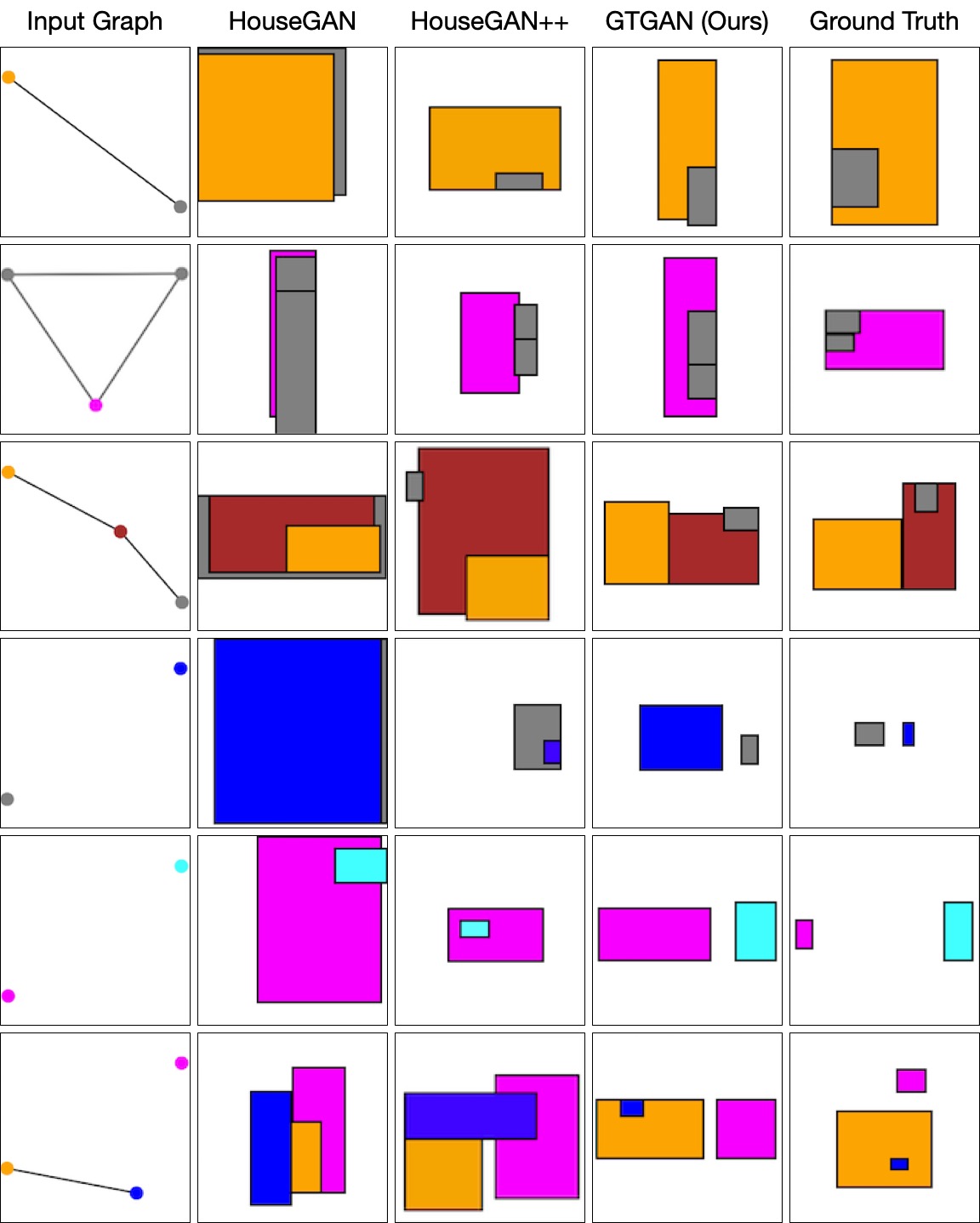}
	\includegraphics[width=0.328\linewidth]{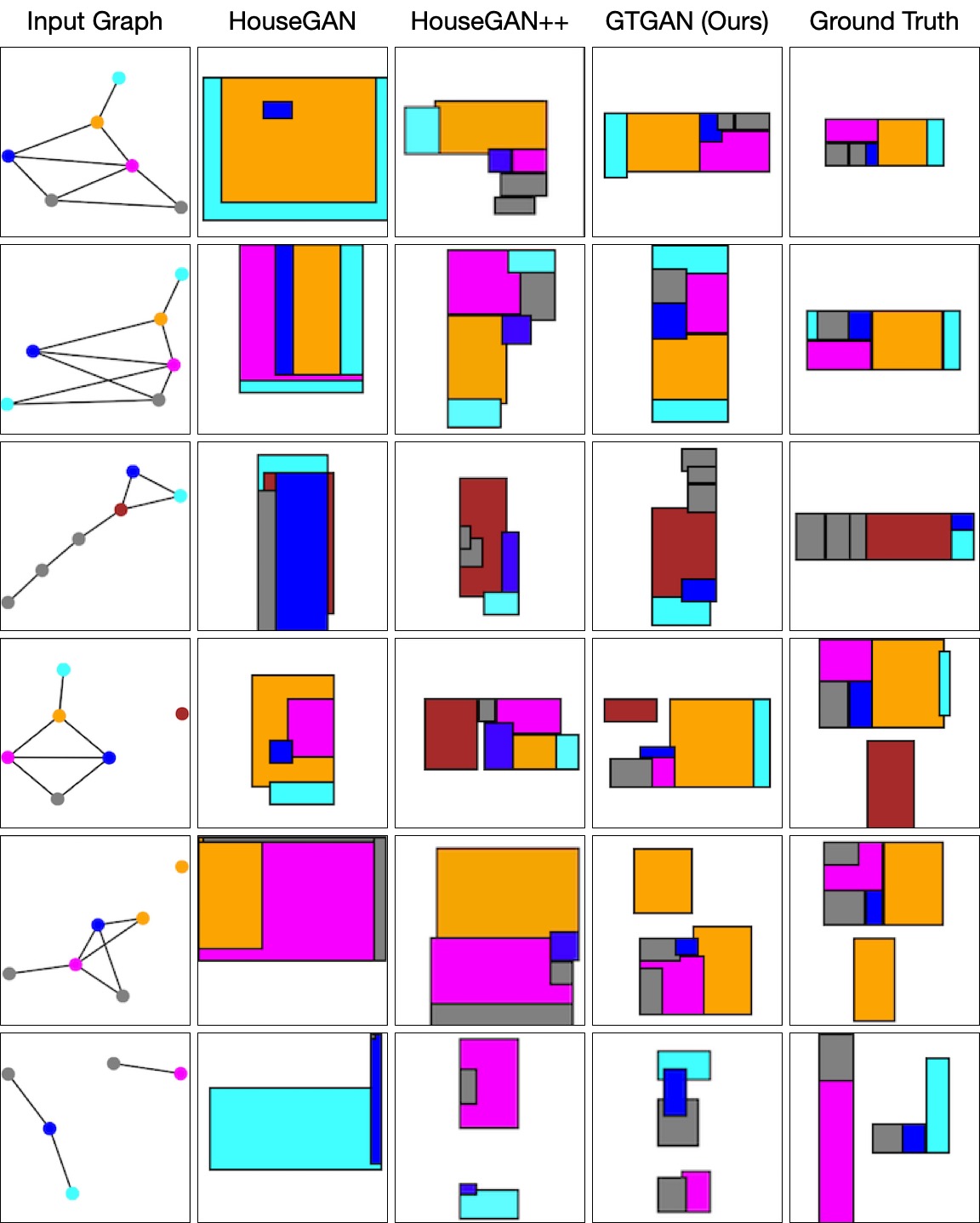}
	\includegraphics[width=0.328\linewidth]{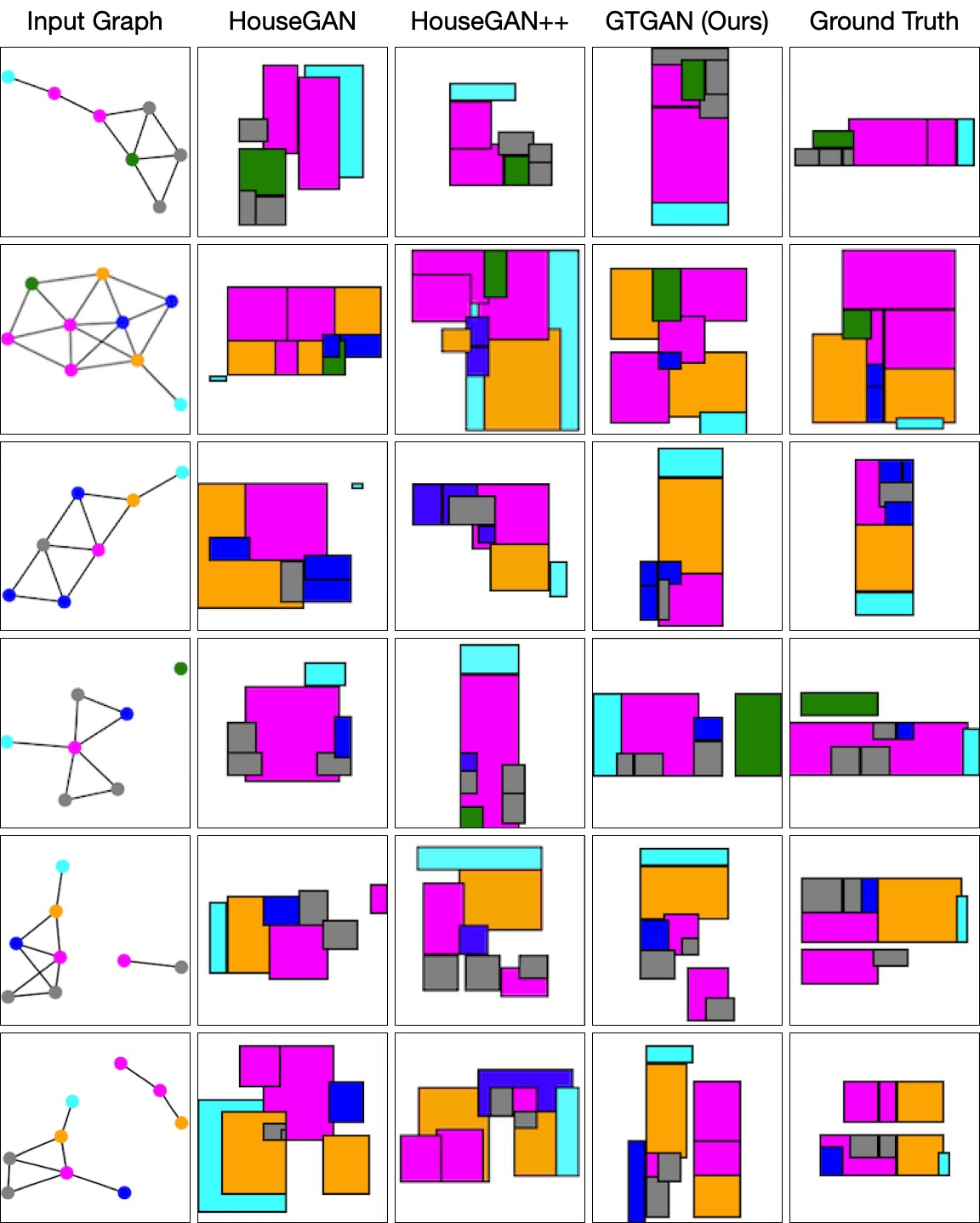}
	\caption{Visualization results compared with HouseGAN~\cite{nauata2020house} and HouseGAN++ \cite{nauata2021house} on ``1-3'' (\textbf{left}), ``4-6'' (\textbf{middle}), and ``7-9'' (\textbf{right}) subset. The last three rows contain non-connected nodes.}
	\label{fig:results_A}
	\vspace{-0.4cm}
\end{figure*}

\begin{figure*}[!t] \small
	\centering
	\includegraphics[width=0.35\linewidth]{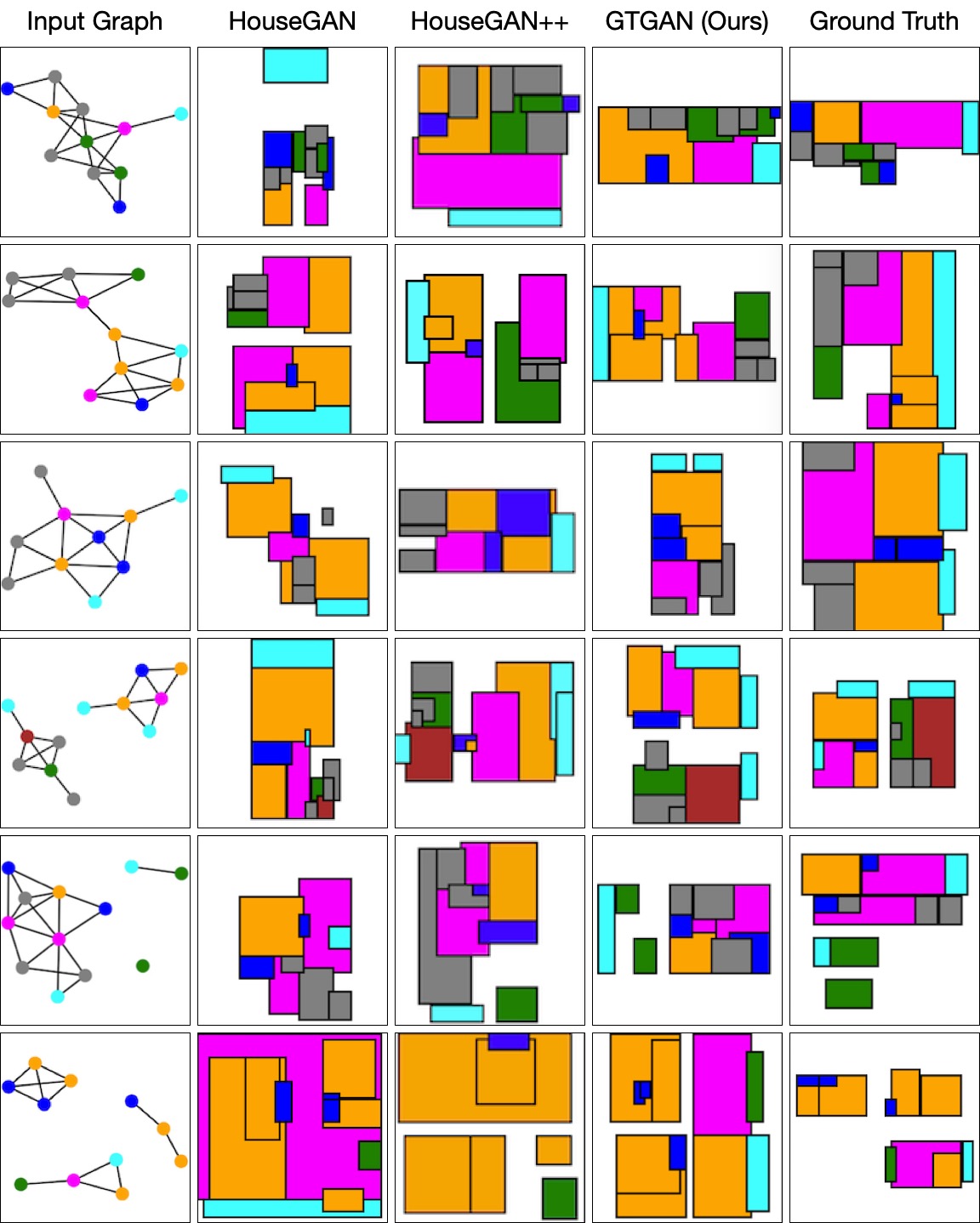}
	\includegraphics[width=0.35\linewidth]{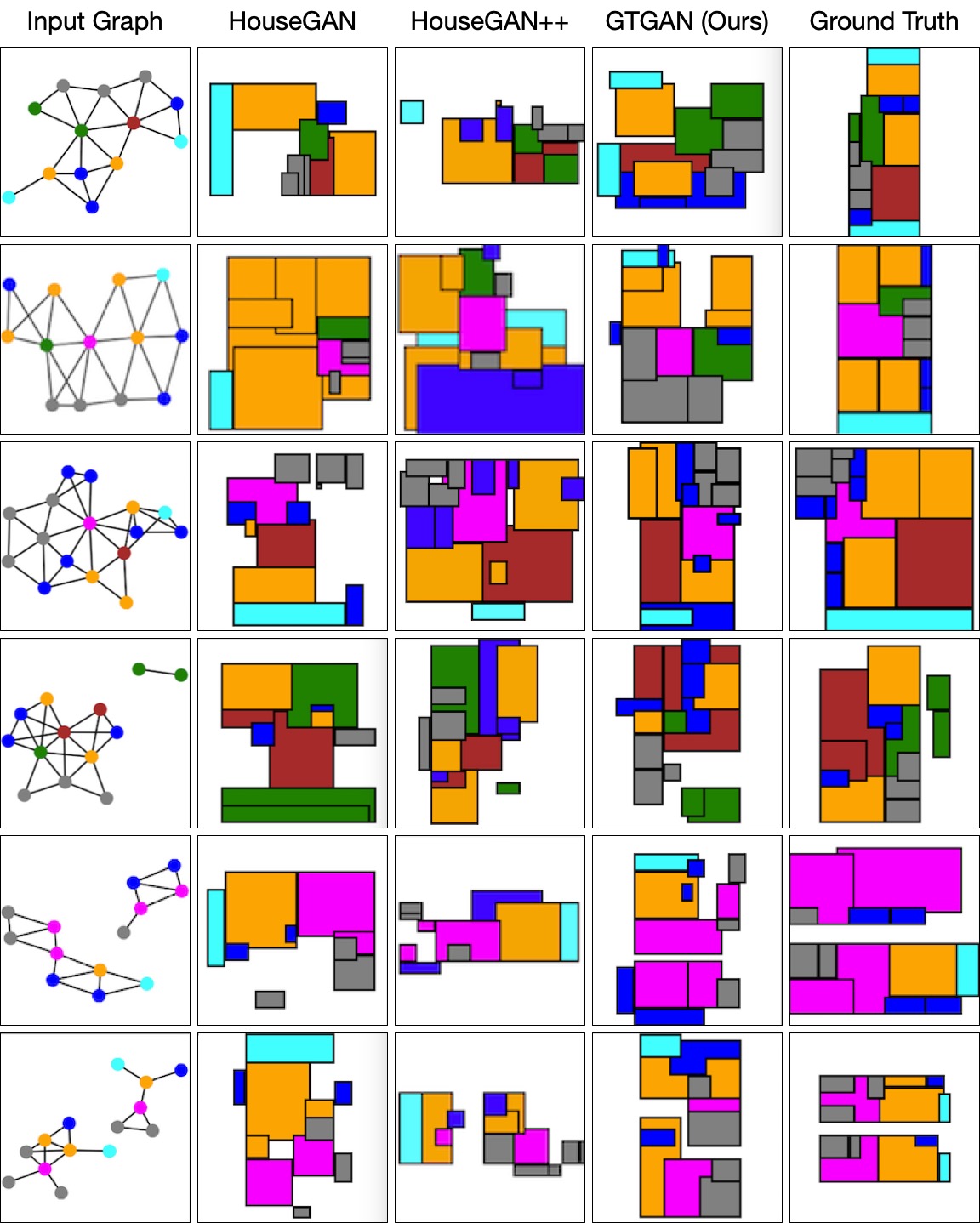}
	\caption{Visualization results compared with HouseGAN~\cite{nauata2020house} and HouseGAN++ \cite{nauata2021house} on ``10-12'' (\textbf{left}) and ``13+'' (\textbf{right}) subset. The last three samples contain non-connected nodes.}
	\label{fig:results_D}
	\vspace{-0.4cm}
\end{figure*}

\subsection{Results on House Layout Generation}

\noindent \textbf{Datasets.}
We follow HouseGAN \cite{nauata2020house} and conduct house layout generation experiments on the LIFULL HOME's dataset, which have different rooms, \ie, living room, kitchen, bedroom, bathroom, closet, balcony, corridor, dining room, laundry room, and unknown.

\noindent \textbf{Evaluation Metrics.}
We follow HouseGAN~\cite{nauata2020house} and adopt realism, diversity, and compatibility metrics to evaluate the performance of the proposed method.
Specifically, we follow~\cite{nauata2020house} and divide the training samples into five subsets based on the number of rooms, \ie, 1-3, 4-6, 7-9, 10-12, and 13+. 
When generating layouts in a subset, we train the models while excluding samples in the same subset so that they cannot simply memorize. 
1) We use the average user rating (12 Ph.D. students and 10 professional architects) to measure realism. 
We provide 75 generated results with ground truths or 75 results generated by another method for comparison.
A subject can give four ratings, \ie, better (+1), equally-good (+1), worse (-1), or equally-bad (-1).
2) The Fr\'echet inception distance (FID) \cite{heusel2017gans} measures the diversity with the rasterized layout images.
We rasterize a layout by setting the background to white and then sorting the rooms in decreasing order of area, finally painting each room with a color based on its room type. We follow \cite{nauata2020house} and use 5,000 samples to compute the FID metric.
3) The compatibility of the bubble diagram is determined by the graph editing distance \cite{abu2015exact} between the input bubble diagram and the bubble diagram constructed from the generated layout.

\begin{figure*}[!t] \small
	\centering
	\includegraphics[width=0.95\linewidth]{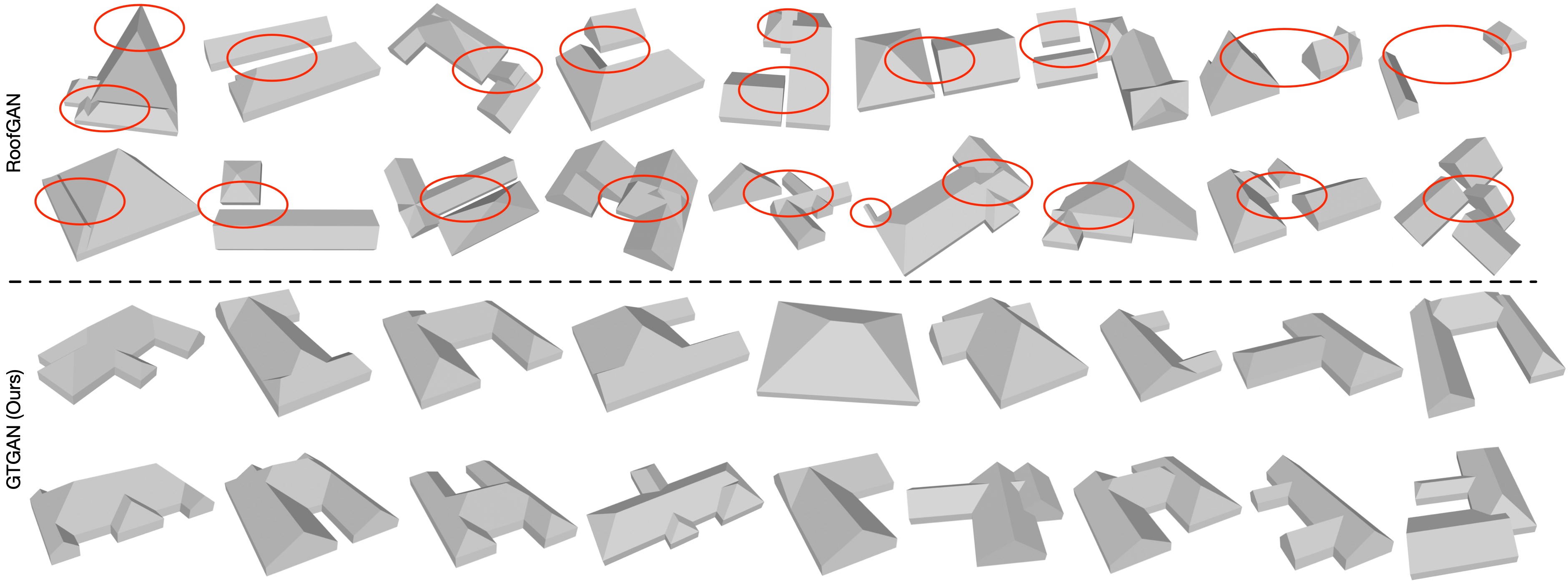}
	\caption{Visualization results compared with the proposed GTGAN (\textbf{bottom two rows}) and RoofGAN~(\textbf{top two rows}). We see that GTGAN can generate more realistic roof structures than RoofGAN. Red ovals highlight non-realistic roof structures generated by RoofGAN.}
	\label{fig:results_roof}
	\vspace{-0.4cm}
\end{figure*}

\begin{table*}[!t] \small
	\centering
	\resizebox{0.8\linewidth}{!}{%
		\begin{tabular}{lccccccccccc} \toprule
			\multirow{2}{*}{Method}  & Realism $\uparrow$ & \multicolumn{5}{c}{Diversity $\downarrow$} & \multicolumn{5}{c}{Compatibility $\downarrow$} \\ \cmidrule(lr){2-2}  \cmidrule(lr){3-7} \cmidrule(lr){8-12} 
			& All Groups & 1-3 & 4-6   & 7-9 & 10-12  & 13+  & 1-3  & 4-6  & 7-9 & 10-12 & 13+      \\ \hline	
			CNN-only & -0.54 & 13.2 & 26.6 & 43.6 & 54.6 & 90.0 & 0.4 & 3.1 & 8.1 & 15.8 &34.7 \\
			GCN         & 0.14 & 18.6 & 17.0 & 18.1 & 22.7 & 31.5 &  {\color{orange}0.1} & {\color{orange}0.8} &  {\color{cyan} 2.3} & {\color{cyan} 3.2} & {\color{cyan} 3.7}\\
			Ashual et al.~\cite{ashual2019specifying} & -0.55 & 64.0 & 92.2& 87.6& 122.8& 149.9&  0.2 & 2.7& 6.2& 19.2& 36.0\\
			Johnson et al.~\cite{johnson2018image}  & -0.58 & 69.8& 86.9& 80.1& 117.5& 123.2& 0.2 & 2.6& 5.2& 17.5& 29.3\\
			HouseGAN~\cite{nauata2020house}       & {\color{orange}0.17} & {\color{orange}13.6}&  {\color{orange}9.4} & {\color{orange}14.4}&  {\color{orange}11.6} & {\color{orange}20.1} & {\color{orange}0.1} &  1.1 & 2.9&  3.9 & 10.8 \\
			HouseGAN++$\ast$ \cite{nauata2021house} & {\color{cyan}0.19} & {\color{cyan} 11.8} & {\color{cyan} 7.6} & {\color{cyan} 12.2} & {\color{cyan} 10.1} & {\color{cyan} 18.3} & {\color{cyan} 0.08} & {\color{cyan} 0.77} &  {\color{orange} 2.52} & {\color{orange}3.65} & {\color{orange}7.43} \\
			GTGAN (Ours)                                 & {\color{blue} 0.25} & {\color{blue} 7.1}  & {\color{blue} 5.4}   & {\color{blue} 9.6}  & {\color{blue} 7.5} & {\color{blue} 16.9}  &  {\color{blue} 0.06} & {\color{blue} 0.62} & {\color{blue} 2.14}  & {\color{blue} 2.63}  & {\color{blue} 3.42} \\
			\bottomrule	
	\end{tabular}}
	\caption{Quantitative results of house layout generation on the LIFULL HOME's dataset. The colors {\color{blue} blue}, {\color{cyan} cyan}, and {\color{orange} orange} represent the first, the second, and third best results, respectively. HouseGAN++$\ast$ \cite{nauata2021house} is reproduced by us.}
	\label{tab:house_reuslts}
	\vspace{-0.4cm}
\end{table*}

\noindent \textbf{Quantitative Comparisons.}
To evaluate the effectiveness of GTGAN on house layout generation, we compare it with several leading methods, \ie, CNN-only, GCN, Ashual et al.~\cite{ashual2019specifying}, Johnson et al.~\cite{johnson2018image}, HouseGAN~\cite{nauata2020house}, and HouseGAN++ \cite{nauata2021house}.
We follow the same setups in \cite{nauata2020house} to reproduce the results of Ashual et al.~\cite{ashual2019specifying} and Johnson et al.~\cite{johnson2018image} for fair comparisons.
Table \ref{tab:house_reuslts} shows our main results on the five subsets. 
Note that we train HouseGAN++ \cite{nauata2021house} on this dataset using the public source code for a fair comparison, which is denoted as HouseGAN++$\ast$. 
We observe that GTGAN outperforms other competing methods in all the metrics, validating the effectiveness of GTGAN.

\noindent \textbf{Qualitative Comparisons.}
We compare GTGAN with the most related methods, \ie, HouseGAN~\cite{nauata2020house} and HouseGAN++ \cite{nauata2021house}.
The visualization results on the five subsets are shown in Figures~\ref{fig:results_A} and \ref{fig:results_D}. 
It is easy to tell that GTGAN generates more realistic and reasonable house layouts than the leading methods, \ie, HouseGAN and HouseGAN++.
For instance, HouseGAN generates improper room sizes or shapes for certain room types, \eg, the closet, the kitchen, and the closet in the last three rows of Figure~\ref{fig:results_A}(left), respectively, are too big. 
Moreover, HouseGAN generates misalignment of rooms, \eg, the balcony, the kitchen, and the living room in the first three samples of Figure~\ref{fig:results_A}(middle) do not align well with other rooms.
Lastly, both HouseGAN and  HouseGAN++ cannot generate non-connected rooms well, \eg, the corridor, the bathroom, and the bedroom in the last three samples of Figure~\ref{fig:results_A}(right) are not accurately generated because they are not connected to other rooms.
In contrast, the proposed method alleviates all three problems to a certain extent and generates more realistic and reasonable house layouts.

\subsection{Results on House Roof Generation}


\begin{figure*}[htbp]
	\centering
	\begin{minipage}[b]{0.305\linewidth}
		\centering
		\includegraphics[width=\textwidth]{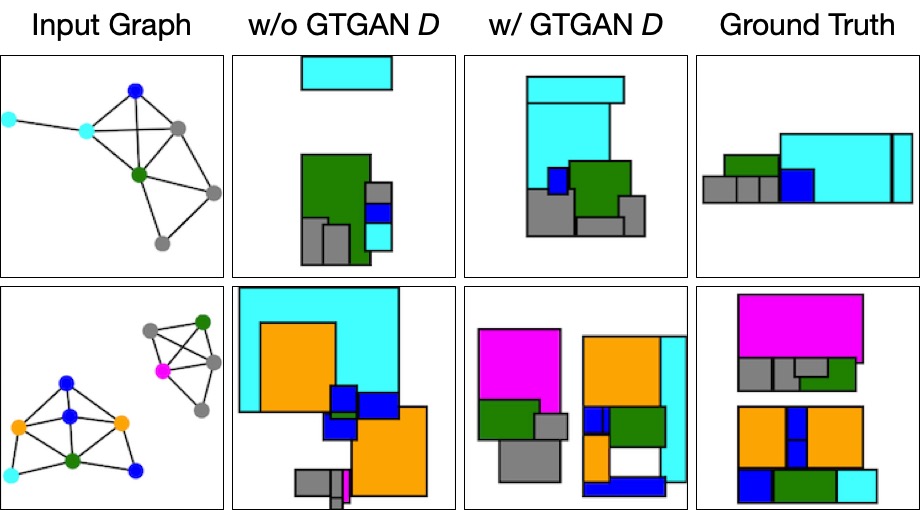}
		\caption{Comparison between w/o and w/ our GTGAN $D$.}
		\label{fig:node_discriminator}
		\vspace{-0.2cm}
	\end{minipage}
	\,
	\begin{minipage}[b]{0.325\linewidth}
		\centering
		\includegraphics[width=\textwidth]{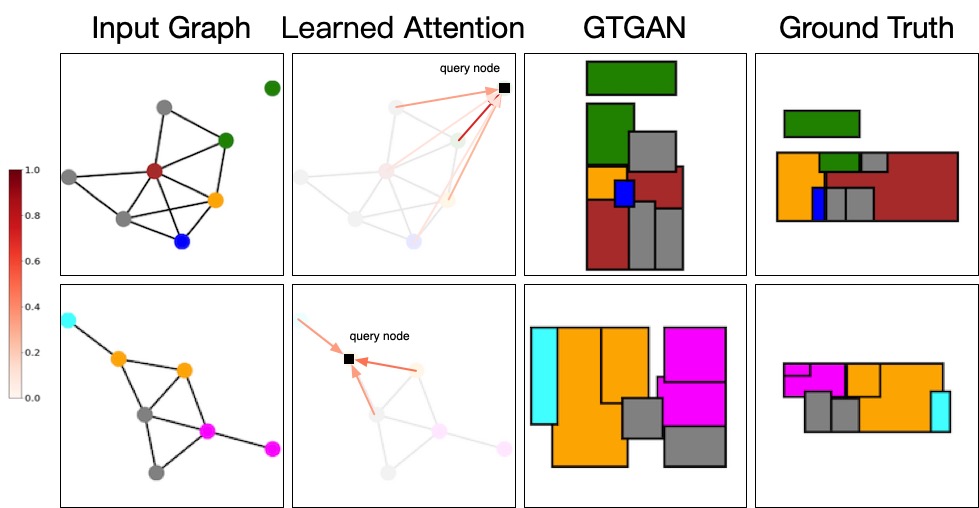}
		\caption{Visualization of the learned node attention.}
		\label{fig:node_attention}
		\vspace{-0.2cm}
	\end{minipage}
	\,
	\begin{minipage}[b]{0.34\linewidth}
		\centering
		\resizebox{1\linewidth}{!}{%
			\begin{tabular}{lcccc} \toprule
				\multirow{2}{*}{Method} & \multicolumn{2}{c}{3 Primitives} & \multicolumn{2}{c}{4 Primitives} \\ \cmidrule(lr){2-3} \cmidrule(lr){4-5}
				& FID $\downarrow$ & RMMD $\downarrow$	& FID $\downarrow$ & RMMD $\downarrow$ \\ \midrule
				PQ-Net \cite{wu2020pq} & {\color{orange} 13.0} & 10.4 & {\color{orange} 14.6} & 12.9 \\
				HouseGAN \cite{nauata2020house} & 27.5 & {\color{orange} 8.5} & 27.2 & {\color{orange} 12.5} \\
				RoofGAN \cite{qian2021roof} & {\color{cyan}11.1} & {\color{cyan}7.5} & {\color{cyan}13.8} & {\color{cyan}10.9} \\ 
				GTGAN (Ours) & {\color{blue}9.3} & {\color{blue}5.5} & {\color{blue}9.6} & {\color{blue}7.2}\\ \bottomrule
		\end{tabular}}
		\captionof{table}{Quantitative results of house roof generation on the CAD-style roof geometry dataset.}
		\label{tab:roof_reuslts}
		\vspace{-0.2cm}
	\end{minipage}
\end{figure*}

\noindent \textbf{Datasets and Evaluation Metrics.}
We follow RoofGAN \cite{qian2021roof} and  conduct extensive experiments on the CAD-style roof geometry dataset proposed in \cite{qian2021roof}. We follow \cite{qian2021roof} and use the FID \cite{heusel2017gans} and the minimum matching distance (RMMD) as the evaluation metrics for a fair comparison.


\noindent \textbf{Quantitative Comparisons.}
To evaluate the effectiveness of GTGAN on house roof generation, we compare it with three leading methods, \ie, PQ-Net \cite{wu2020pq}, HouseGAN \cite{nauata2020house}, and RoofGAN \cite{qian2021roof}.
Table \ref{tab:roof_reuslts} shows the comparison results on both 3 and 4 primitives.
When generating roofs, we follow \cite{qian2021roof} and split the training and test sets based on
the number of primitives to prevent simply copying and pasting.
We observe that GTGAN outperforms the other three competing methods in both metrics,  demonstrating the effectiveness of our method.

\noindent \textbf{Qualitative Comparisons.}
We compare GTGAN with the most related method, \ie, RoofGAN \cite{nauata2020house}.
The visualization results are shown in Figure \ref{fig:results_roof}. 
Clearly, we observe that GTGAN generates more realistic and reasonable roof structures than the leading method RoofGAN.
For instance, RoofGAN generates isolated, too-long, too-high, or too-thin roofs. 
It also produces poor relationships between different components, which results in unrealistic polygonal shapes and topology.
In contrast, GTGAN alleviates all these problems to a certain extent and generates more complex and realistic combinations of roof primitives. 
Moreover, GTGAN also produces more diverse roofs, which is another advantage of our method.

\subsection{Ablation Study}
We conduct extensive ablation studies on house layout generation (``10-12'' subset) to evaluate the effectiveness of each component of the proposed GTGAN.

\noindent \textbf{Baselines Models.}
GTGAN has 11 baselines, as shown in Table~\ref{tab:abla}: 
(1) B1 is our baseline combining HouseGAN $G$ and HouseGAN $D$ (\ie, the original HouseGAN~\cite{nauata2020house}).
(2) B2 adopts the combination of GTGAN $G$ and HouseGAN $D$.
(3) B3 combines HouseGAN $G$ and GTGAN $D$. 
(4) B4 employs both GTGAN $G$ and GTGAN $D$.
(5) B5 is our baseline without using NNA.
(6) B6 is our baseline without using CNA.
(7) B7 is our baseline without using GMB.
(8) B8 is our baseline using Transformer layers instead of Conv-MPN layers.
(9) B9 is the variation using Eq.~\eqref{eq:eq1} instead of Eq.~\eqref{eq:eq}.
(10) B10 is the variation using Eq.~\eqref{eq:eq2}  instead of Eq.~\eqref{eq:eq}.
(11) B11 is our full model, using the proposed graph-based cycle-consistency loss upon B4.

\begin{table}[!t] \small
	\centering
	\resizebox{1\linewidth}{!}{%
		\begin{tabular}{ccccccc} \toprule
			\# & Generator    & Discriminator  & FID $\downarrow$ & Compatibility  $\downarrow$ \\ \midrule	
			B1 & HouseGAN  \cite{nauata2020house}  & HouseGAN  \cite{nauata2020house} & 11.6  & 3.90 \\ 
			B2 & GTGAN & HouseGAN  \cite{nauata2020house} & 10.3 & 3.49 \\
			B3 & HouseGAN   \cite{nauata2020house} & GTGAN & 9.5  & 3.22 \\
			B4 & GTGAN & GTGAN & 8.2 &  2.95 \\  \hline
			B5 & GTGAN w/o NNA & GTGAN & 9.4 & 3.46 \\
			B6 & GTGAN w/o CNA & GTGAN & 9.7 & 3.68 \\
			B7 & GTGAN w/o GMB & GTGAN &  9.5 & 3.61 \\
			B8 & GTGAN w/ Transformer Layers & GTGAN & 8.9 & 3.45 \\ \hline
			B9 & GTGAN w/ Eq.~\eqref{eq:eq1} & GTGAN & 9.4 & 3.29 \\ 
			B10 & GTGAN w/ Eq.~\eqref{eq:eq2} & GTGAN & 9.8 & 3.32 \\ \hline
			B11  &  \multicolumn{2}{c}{B4 + Graph-Based Cycle-Consistency Loss}  & \textbf{7.5} &  \textbf{2.63} \\ \bottomrule
	\end{tabular}}
	\caption{Ablation study of GTGAN on house layout generation.}
	\label{tab:abla}
	\vspace{-0.4cm}
\end{table}

\noindent \textbf{Ablation Analysis.}
The results of the ablation study, shown in Table~\ref{tab:abla}, prove that our graph Transformer generator $G$ and node classification-based discriminator $D$ improve the generation performance over the baseline models, validating the effectiveness of the proposed framework.
Specifically, when using our generator, B2 yields further improvements over B1, meaning that our generator learns local and global relations across connected and non-connected nodes more effectively, confirming our design motivation.
B3 outperforms B1, demonstrating the importance of using our discriminator to generate semantically consistent rooms according to the input graph nodes.
We show the comparison results in Figure \ref{fig:node_discriminator}.
We see that using our discriminator is helping to preserve the room information in the generated floorplans, leading to a better house layout.
Moreover, we observe that B4 generates significantly better results than B2 and B3, further confirming our network design.

We also observe that not using NNA or CNA significantly reduces performance in B5 or B6, which validates the effectiveness of both NNA and CNA.
Also, without using the proposed GMB in B7, the performance drops a lot on both metrics.
Meanwhile, using Transformer layers (B8) other than Conv-MPN layers slightly reduces the performance, which means that using a mixed model of CNN and Transformer can achieve better results.
When we use Eq.~\eqref{eq:eq1} and Eq.~\eqref{eq:eq2} instead of Eq.~\eqref{eq:eq} in B9 and B10, the performance drops slightly, which also proves the rationality of our model design.
Lastly,  our full model, B11, significantly outperforms B4, clearly demonstrating
the effectiveness of the proposed graph-based cycle-consistency loss.

\noindent \textbf{Visulization of Attention Weights.} We show two examples of the learned node attention in Figure~\ref{fig:node_attention}. The query nodes are colored in black, whereas the edges are colored according to the magnitude of the attention weights, which can be referred to by the color bar on the left. We can observe that the proposed method indeed learns the relationships between nodes.

\section{Conclusion}
With this work, we are the first to explore using a Transformer-based architecture for the graph-constrained house generation task. 
We provide three contributions, \ie, a graph Transformer-based generator, a node classification-based discriminator, and a graph-based cycle-consistency loss.
The first is employed to model local and global relations across connected and non-connected nodes in a graph.
The second component is used to preserve high-level, semantically discriminative features for different house components.
The last is used to preserve relative spatial relationships between ground truth and generated graphs.
Extensive experiments in terms of both human and automatic evaluation demonstrate that GTGAN achieves remarkably better performance than existing approaches on both house layout and roof generation tasks.  

\noindent\textbf{Acknowledgments.} This work was partly supported by the ETH Zurich General Fund (OK), the Alexander von Humboldt Foundation, and the EU H2020 project AI4Media (No. 951911).

\clearpage
{\small
\bibliographystyle{ieee_fullname}
\bibliography{egbib}
}


\end{document}